\documentclass{article}
\usepackage{url}
\usepackage{graphicx}
\usepackage{authblk}
\usepackage{hyperref}
\begin{document}
\title{GraphSL: An Open-Source Library for Graph Source Localization Approaches and Benchmark Datasets}
\author{Junxiang Wang}
\author{Liang Zhao}
\affil{Emory University}
\date{}
\maketitle
\section*{Summary}

We introduce GraphSL, a new library for studying the graph source localization problem. graph diffusion and graph source localization are inverse problems in nature: graph diffusion predicts information diffusions from information sources, while graph source localization predicts information sources from information diffusions. GraphSL facilitates the exploration of various graph diffusion models for simulating information diffusions and enables the evaluation of cutting-edge source localization approaches on established benchmark datasets. The source code of GraphSL is made available at Github Repository (\url{https://github.com/xianggebenben/GraphSL}). Bug reports and feedback can be directed to the Github issues page (\url{https://github.com/xianggebenben/GraphSL/issues}).

\section*{Statement of Need}
\begin{figure}
    \centering
    \includegraphics[width=\linewidth]{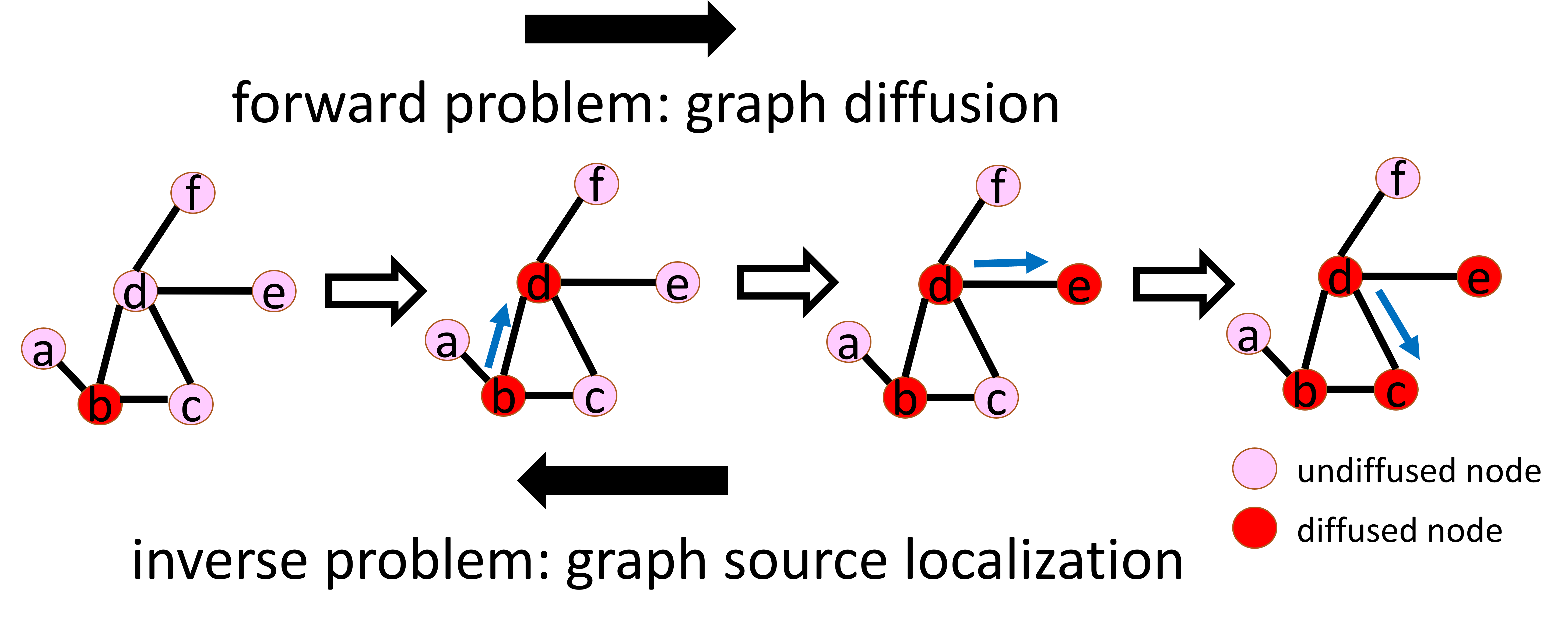}
    \caption{An example of graph source localization}
\label{fig:example}
\end{figure}

Graph diffusion is a fundamental task in graph learning, which aims to predict future information diffusions given information sources. Its inverse problem is graph source localization, which is an extremely important topic even though rarely explored: it focuses on the detection of information sources given their future information diffusions. As illustrated in \autoref{fig:example}, graph diffusion seeks to predict the information diffusion $\{b,c,d,e\}$ from a source node $b$, whereas graph source localization aims to identify the source node $b$ from the information diffusion $\{b,c,d,e\}$. Graph source localization spans a broad spectrum of promising research and real-world applications such as rumor detection \cite{gallotti2020assessing}, tracking of sources for computer viruses\cite{kephart1993measuring}, and failure detection in smart grids \cite{amin2007preventing}. Please refer to the survey paper \cite{jiang2016identifying} for more information. Hence, the graph source localization problem demands attention and extensive investigations from machine learning researchers.

Due to its importance, some open-source tools have been developed to support research of the graph source localization problem. Two recent examples are cosasi \cite{McCabe2022joss} and RPaSDT \cite{frkaszczak2022rpasdt}. However, they do not support various simulations of information diffusion, and they also miss real-world benchmark datasets and state-of-the-art source localization approaches. To fill this gap, we propose a new library GraphSL: the first one to include real-world benchmark datasets and recent source localization methods to our knowledge, enabling researchers and practitioners to evaluate novel techniques against appropriate baselines easily. These methods do not require prior assumptions about the source (e.g. single source or multiple sources) and can handle graph source localization based on various diffusion simulation models such as Independent Cascade (IC) and Linear Threshold (LT) \cite{shakarian2015independent}. Our GraphSL library is standardized: for instance, tests of all source inference methods return a Metric object, which provides five performance metrics (accuracy, precision, recall, F-score, and area under ROC curve) for performance evaluation.

Our GraphSL library targets both developers and practical users: they are free to add algorithms and datasets for personal needs by following the guidelines in the "Contact" section of README.md(\url{https://github.com/xianggebenben/GraphSL/blob/main/README.md}).

\begin{figure}
    \centering
    \includegraphics[width=\linewidth]{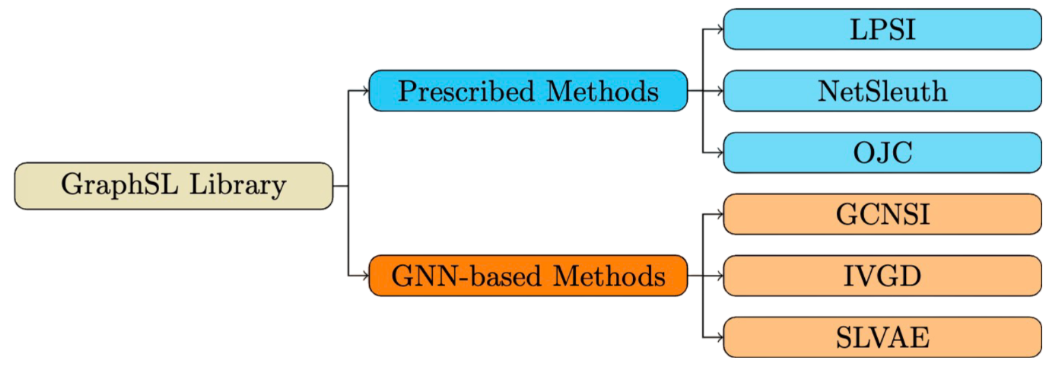}
    \caption{The hierarchical structure of the GraphSL library: in total six algorithms are implemented, which can be divided into two categories: prescribed methods that rely on hand-crafted rules and GNN-based methods which learn rules from graph data.}
    \label{fig:overview}
\end{figure}
\section*{Methods and Benchmark Datasets}

The structure of our GraphSL library is depicted in \autoref{fig:overview}. Existing methods can be categorized into two groups: Prescribed methods and Graph Neural Networks (GNN)-based methods.

Prescribed methods rely on hand-crafted rules and heuristics. For instance, LPSI assumes that nodes surrounded by larger proportions of infected nodes are more likely to be source nodes \cite{wang2017multiple}. NetSleuth employs the Minimum Description Length principle to identify the optimal set of source nodes and virus propagation ripple \cite{prakash2012spotting}. OJC identifies a set of nodes (Jordan cover) that cover all observed infected nodes with the minimum radius \cite{zhu2017catch}.

GNN-based methods learn rules from graph data in an end-to-end manner by capturing graph topology and neighboring information. For example, GCNSI utilizes LPSI to enhance input and then applies Graph Convolutional Networks (GCN) for source identification \cite{dong2019multiple}; IVGD introduces a graph residual scenario to make existing graph diffusion models invertible, and it devises a new set of validity-aware layers to project inferred sources to feasible regions \cite{IVGD_www22}. SLVAE uses forward diffusion estimation and deep generative models to approximate source distribution, leveraging prior knowledge for generalization under arbitrary diffusion patterns \cite{ling2022source}.
\begin{table}[!ht]
    \centering
    \begin{tabular}{c|c|c}
    \hline\hline
        Dataset & \#Node & \#Edge \\ \hline
        Karate  \cite{lusseau2003bottlenose} & 34 & 78 \\ \hline
        Dolphins \cite{lusseau2003bottlenose} & 62 & 159 \\ \hline
        Jazz \cite{gleiser2003community} & 198 & 2,742\\ \hline
        Network   Science \cite{newman2006finding} & 1,589 & 2,742 \\ \hline
        Cora-ML \cite{mccallum2000automating} & 2,810 & 7,981\\ \hline
        Power   Grid \cite{watts1998collective} & 4,941 & 6,594\\ \hline\hline
    \end{tabular}
    \caption{ Six benchmark graph datasets: their numbers of nodes and edges.}
    \label{tab:statistics} 
\end{table}
\\
Aside from methods, we also release six benchmark graph datasets to facilitate the research of graph source localization, whose statistics are shown in \autoref{tab:statistics}. Information sources and diffusions can be generated by the function diffusion\_generation (\url{https://graphsl.readthedocs.io/en/latest/GraphSL.html#GraphSL.utils.diffusion_generation}).

\section*{Availability and Documentation}

GraphSL is available under the MIT License. The library may be cloned from the GitHub repository (\url{https://github.com/xianggebenben/GraphSL}), or can be installed by pip: pip install GraphSL. Documentation is provided via Read the Docs (\url{https://graphsl.readthedocs.io/en/latest/index.html}), including a quickstart introducing major functionality and a detailed API reference. Extensive unit testing is employed throughout the library. 
\bibliographystyle{plain}
\bibliography{ref}
\end{document}